\def\year{2020}\relax
\documentclass[letterpaper]{article} 
\usepackage[dvipsnames]{xcolor}
\usepackage{aaai20}  
\usepackage{times}  
\usepackage{helvet} 
\usepackage{courier}  
\usepackage[hyphens]{url}  
\usepackage{graphicx} 
\usepackage{graphicx}  
\usepackage{graphicx,parskip,setspace,tabularx,xspace,array,multirow,listings,amsmath,adjustbox,threeparttable,subcaption,makecell} 
\usepackage[justification=justified]{caption}
\usepackage{float}
\usepackage{hhline}

\title{Going Beneath the Surface: Evaluating Image Captioning for Grammaticality, Truthfulness and Diversity}

\author{\Large \textbf{Huiyuan Xie\textsuperscript{\rm 1}, Tom Sherborne\textsuperscript{\rm 2}, Alexander Kuhnle\textsuperscript{\rm 1}, Ann Copestake\textsuperscript{\rm 1}} \\
\textsuperscript{\rm 1}Department of Computer Science and Technology, University of Cambridge \\
\textsuperscript{\rm 2}School of Informatics, University of Edinburgh \\
\{hx255,aok25,aac10\}@cam.ac.uk, tom.sherborne@ed.ac.uk
}

\begin{document}

\maketitle

\begin{abstract}
Image captioning as a multimodal task has drawn much interest in recent years. However, evaluation for this task remains a challenging problem. Existing evaluation metrics focus on surface similarity between a candidate caption and a set of reference captions, and do not check the actual relation between a caption and the underlying visual content. We introduce a new diagnostic evaluation framework for the task of image captioning, with the goal of directly assessing models for grammaticality, truthfulness and diversity (GTD) of generated captions. We demonstrate the potential of our evaluation framework by evaluating existing image captioning models on a wide ranging set of synthetic datasets that we construct for diagnostic evaluation. We empirically show how the GTD evaluation framework, in combination with diagnostic datasets, can provide insights into model capabilities and limitations to supplement standard evaluations.
\end{abstract}

\section{Introduction}
Automatically generating text to describe the content of images, also known as image captioning, is a multimodal task of considerable interest in both the computer vision and the NLP communities. Image captioning can be framed as a translation task from an image to a descriptive natural language statement. Many existing captioning models \cite{vinyals2015show,donahue2015long,yao2017boosting,aneja2018convolutional} follow the typical encoder-decoder framework where a convolutional network is used to condense images into visual feature representations, combined with a recurrent network for language generation. While these models demonstrate promising results, quantifying image captioning performance remains a challenging problem, in a similar way to other generative tasks \cite{radev2003evaluation,gatt2018survey}.

\begin{figure}[t]
\centering
\begin{minipage}{0.35\linewidth}
\includegraphics[width=\linewidth]{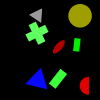}
\end{minipage}%
\hspace{0.03\linewidth}%
\begin{minipage}{0.55\linewidth}
\small
\textit{Caption 1}: \textcolor{RoyalBlue}{A circle is above a green rectangle.} \\[0.1cm]
\textit{Caption 2}: \textcolor{RoyalBlue}{A blue triangle is to the left of a semicircle.} \\[0.1cm]
\textit{Caption 3}: \textcolor{RoyalBlue}{A semicircle is below a gray triangle.} \\[0.1cm]
\textit{Caption 4}: \textcolor{BrickRed}{A semicircle is to the left of a triangle.}
\end{minipage}
\caption{ShapeWorld example: spatial statements in the context of multiple shapes. The first three statements are truthful and diverse descriptions of the image. The fourth statement is wrong, but nonetheless exhibits a high degree of n-gram overlap with the true reference captions.}
\label{fig:shapeworld}
\end{figure}

Evaluating candidate captions for human preference is slow and laborious. To alleviate this problem, many automatic evaluation metrics have been proposed, such as BLEU \cite{papineni2002bleu}, METEOR \cite{banerjee2005meteor}, ROUGE \cite{lin2004rouge} and CIDEr \cite{vedantam2015cider}. These n-gram-based metrics evaluate captioning performance based on surface similarity between a candidate caption and reference statements. A more recent evaluation metric for image captioning is SPICE \cite{anderson2016spice}, which takes into account semantic propositional content of generated captions by scoring a caption based upon a graph-based semantic representation transformed from reference captions.

The rationale behind these evaluation metrics is that human reference captions serve as an approximate target and comparing model outputs to this target is a proxy for how well a system performs. Thus, a candidate caption is not directly evaluated with respect to image content, but compared to a set of human statements about that image. 

However, in image captioning, visual scenes with multiple objects and relations correspond to a diversity of valid descriptions. Consider the example image and captions from the ShapeWorld framework \cite{kuhnle2017shapeworld} shown in Figure \ref{fig:shapeworld}. The first three captions are true statements about the image and express relevant ideas, but describe different objects, attributes and spatial relationships, while the fourth caption is wrong despite referring to the same objects as in the third caption. This casts doubt on the sufficiency of using a set of reference captions to approximate the content of an image. We argue that, while existing metrics have undeniably been useful for real-world captioning evaluation, their focus on approximate surface comparison limits deeper insights into the learning process and eventual behavior of captioning models.

To address this problem, we propose a set of principled evaluation criteria which evaluate image captioning models for grammaticality, truthfulness and diversity (GTD). These criteria correspond to necessary requirements for image captioning systems: (a) that the output is grammatical, (b) that the output statement is true with respect to the image, and (c) that outputs are diverse and mirror the variability of training captions. 

Practical evaluation of GTD is currently only possible on synthetic data. We construct a range of datasets designed for image captioning evaluation. We call this diagnostic evaluation benchmark ShapeWorldICE (\textit{ShapeWorld for Image Captioning Evaluation}). We illustrate the evaluation of specific image captioning models on ShapeWorldICE. We empirically demonstrate that the existing metrics BLEU and SPICE do not capture true caption-image agreement in all scenarios, while the GTD framework allows a fine-grained investigation of how well existing models cope with varied visual situations and linguistic constructions. 

We believe that as a supplementary evaluation method to real-world metrics, the GTD framework provides evaluation insights that are sufficiently interesting to motivate future work. 

\section{Related work}
\subsection{Existing evaluation of image captioning}

As a natural language generation task, image captioning frequently uses evaluation metrics such as BLEU \cite{papineni2002bleu}, METEOR \cite{banerjee2005meteor}, ROUGE \cite{lin2004rouge} and CIDEr \cite{vedantam2015cider}. These metrics use n-gram similarity between the candidate caption and reference captions to approximate the correlation between a candidate caption and the associated ground truth. SPICE \cite{anderson2016spice} is a more recent metric specifically designed for image captioning. For SPICE, both the candidate caption and reference captions are parsed to scene graphs, and the agreement between tuples extracted from these scene graphs is examined. SPICE more closely relates to our truthfulness evaluation than the other metrics, but it still uses overlap comparison to reference captions as a proxy to ground truth. In contrast, our truthfulness metric directly evaluates a candidate caption against a model of the actual visual content.

Many researchers have pointed out problems with existing reference-based metrics including low correlations with human judgment \cite{Elliott2014,anderson2016spice,Kilickaya2017} and strong baselines using nearest-neighbor methods \cite{Devlin2015} or relying solely on object detection \cite{Wang2018}. Fundamental concerns have been raised with respect to BLEU, including variability in parameterization and precise score calculation leading to significantly different results \cite{Post2018}. Its validity as a metric for tasks other than machine translation has been questioned \cite{Reiter2018}, particularly for tasks for which the output content is not narrowly constrained, like dialogue \cite{Liu2016}.

Some recent work focuses on increasing the diversity of generated captions, for which various measures are proposed. Devlin et al. \cite{devlin2015language} explored the concept of caption diversity by evaluating performance on compositionally novel images. van Miltenburg et al \cite{van2018measuring} framed image captioning as a word recall task and proposed several metrics, predominantly focusing on diversity at the word level. However, this direction is still relatively new and lacks standardized benchmarks and metrics.

\subsection{Synthetic datasets}
\label{sec:rw_sw}
Recently, many synthetic datasets have been proposed as diagnostic tools for deep learning models, such as CLEVR \cite{johnson2017clevr} for visual question answering (VQA), the bAbI tasks \cite{weston2015towards} for text understanding and reasoning, and ShapeWorld \cite{kuhnle2017shapeworld} for visually grounded language understanding. The primary motivation is to reduce complexity which is considered irrelevant to the evaluation focus, to enable better control over the data, and to provide more detailed insights into strengths and limitations of existing models.

In this work, we develop the evaluation datasets within the ShapeWorld framework. ShapeWorld is a controlled data generation framework consisting of abstract colored shapes (see Figure \ref{fig:shapeworld} for an example). We use ShapeWorld to generate training and evaluation data for two major reasons. ShapeWorld supports customized data generation according to user specification, which enables a variety of model inspections in terms of language construction, visual complexity and reasoning ability. Another benefit is that each training and test instance generated in ShapeWorld is returned as a triplet of $<$image, caption, world model$>$. The world model stores information about the underlying microworld used to generate an image and a descriptive caption, internally represented as a list of entities with their attributes, such as \textit{shape}, \textit{color}, \textit{position}. During data generation, ShapeWorld randomly samples a world model from a set of available entities and attributes. The generated world model is then used to realize a corresponding instance consisting of image and caption.
The world model gives the actual semantic information contained in an image, which allows evaluation of caption truthfulness.

\begin{table*}[ht]
    \centering
    \begin{tabular}{|c|c|c|c|}
        \hline
        \textbf{Type} & \textbf{Variant} & \textbf{Caption} & \textbf{Image} \\
        \hline
        \multirow{6}{*}{Existential} & 
        \multirow{3}{*}{OneShape} & \textcolor{RoyalBlue}{There is a green cross.} & \multirow[c]{3}{*}{
        \raisebox{-0.05in}{\includegraphics[width=0.4in]{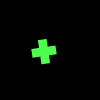}}
        } \\
        & & \textcolor{BrickRed}{A rectangle is green.} & \\
        & & \textcolor{BrickRed}{There is a cyan shape.} & \\
        \cline{2-4}
        & \multirow{3}{*}{MultiShapes} & \textcolor{RoyalBlue}{A shape is a gray triangle.} & \multirow[c]{3}{*}{
        \raisebox{-0.05in}{\includegraphics[width=0.4in]{figs/shapeworld/example3.png}}
        } \\
        & & \textcolor{BrickRed}{There is a square.} & \\
        & & \textcolor{RoyalBlue}{There is a yellow shape.} & \\
        \cline{1-4}
        \multirow{6}{*}{Spatial} & 
        \multirow{3}{*}{TwoShapes} & \textcolor{RoyalBlue}{A square is above a red pentagon.} 
        & \multirow[c]{3}{*}{
        \raisebox{-0.05in}{\includegraphics[width=0.4in]{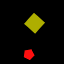}}
        } \\
        & & \textcolor{BrickRed}{A yellow square is above a yellow pentagon.} & \\
        & & \textcolor{BrickRed}{A square is to the left of a pentagon.} & \\
        \cline{2-4}
        & \multirow{3}{*}{MultiShapes} & \textcolor{RoyalBlue}{A blue triangle is to the left of a semicircle.} & \multirow[c]{3}{*}{
        \raisebox{-0.05in}{\includegraphics[width=0.4in]{figs/shapeworld/example3.png}}
        } \\
        & & \textcolor{RoyalBlue}{A circle is above a green rectangle.} & \\
        & & \textcolor{BrickRed}{A semicircle is to the left of a circle.} & \\
        \cline{1-4}
        \multirow{6}{*}{Quantification} & 
        \multirow{3}{*}{Count} & \textcolor{RoyalBlue}{Exactly two rectangles are green.} & \multirow{3}{*}{
        \raisebox{-0.05in}{\includegraphics[width=0.4in]{figs/shapeworld/example3.png}}
        } \\
        & & \textcolor{RoyalBlue}{Exactly one shape is a yellow circle.} & \\
        & & \textcolor{BrickRed}{Exactly zero shapes are ellipses.} & \\
        \cline{2-4}
        & \multirow{3}{*}{Ratio} & \textcolor{RoyalBlue}{A quarter of the shapes are rectangles.} & \multirow[c]{3}{*}{
        \raisebox{-0.05in}{\includegraphics[width=0.4in]{figs/shapeworld/example3.png}}
        } \\
        & & \textcolor{BrickRed}{A third of the rectangles are magenta.} & \\
        & & \textcolor{BrickRed}{Half the shapes are green.} & \\
        \hline
    \end{tabular}
    \caption{Sample captions and images from ShapeWorldICE datasets (truthful captions in blue, false in red). Images from \texttt{Existential-OneShape} only contain one object, while images from \texttt{Spatial-TwoShapes} contain two objects. Images from the other four datasets follow the same distribution with multiple abstract objects present in a visual scene.}
    \label{tab:shapeworld}
\end{table*}

\section{GTD Evaluation Framework}
In the following we introduce GTD in more detail, consider it as an evaluation protocol covering necessary aspects of the multifaceted captioning task, rather than a specific metric.

\subsection{Grammaticality}

An essential criterion for an image captioning model is that the captions generated are grammatically well-formed. Fully accurate assessment of grammaticality in a general context is itself a difficult task, but becomes more feasible in a very constrained context like our diagnostic language data. We take parseability with the English Resource Grammar \cite[ERG]{flickinger2000building} as a surrogate for grammaticality, meaning that a sentence is considered grammatically well-formed if we obtain a parse using the ERG.

The ERG is a broad-coverage grammar based on the head-driven phrase structure grammar (HPSG) framework. It is linguistically precise: sentences only parse if they are valid according to its hand-built rules. It is designed to be general-purpose: verified coverage is around 80\% for Wikipedia, and over 90\% for corpora with shorter sentences and more limited vocabulary (for details see \citeauthor{flickinger2011accuracy}~\shortcite{flickinger2011accuracy}). Since the ShapeWorld training data -- the only language source for models to learn from -- is generated using the same grammar, the ERG has $\sim$100\% coverage of grammaticality in the model output space.

\subsection{Truthfulness}

The second aspect we investigate is truthfulness, that is, whether a candidate caption is compatible with the content of the image it is supposed to describe. We evaluate caption truthfulness on the basis of a linguistically-motivated approach using formal semantics. 
We convert the output of the ERG parse for a grammatical caption to a Dependency Minimal Recursion Semantics (DMRS) graph using the \texttt{pydmrs} tool \cite{copestake2016resources}. Each converted DMRS is a logical semantic graph representation 
 corresponding to the caption. We construct a logical proposition from the DMRS graph, and evaluate it against the actual world model of the corresponding image. A caption can be said to agree with an image only if the proposition evaluates as true on the basis of the world model. By examining the logical agreement between a caption representation and a world model, we can check whether the semantics of this caption agrees with the visual content which the world model represents.  Thus we do not rely on a set of captions as a surrogate for the content of an image, but instead leverage the fact that we have the ground truth, thus enabling the evaluation of true image-caption agreement.

\subsection{Diversity}
While grammaticality and truthfulness are essential requirements for image captions, these criteria alone can easily be ``gamed'' by specializing on a small set of generic statements which are true most of the time. In the context of abstract shapes, such captions include examples like \textit{``There is a shape''} or \textit{``At least zero shapes are blue''} (which is technically true even if there is no blue shape). This motivates the third fundamental requirement of captioning output to be diverse.

As ShapeWorldICE exploits a limited size of open-class words, we emphasize the diversity in ShapeWorldICE at the sentence level rather than the word level. Since the ground-truth reference captions in ShapeWorld are randomly sampled, we take the sampled captions accompanying the test images as a proxy for optimal caption diversity, and compare it with the empirical output diversity of the evaluated model on these test images. Practically, we look at language constructions used and compute the corresponding diversity score as the ratio of observed number versus optimal number:
\[
\text{diversity} = \frac{\#\{\text{model-generated}\}}{\#\{\text{ShapeWorld-generated}\}}
\]
Language constructions here correspond to reduced caption representations which only record whether an object is described by shape (e.g., \textit{``square''}), color (e.g., \textit{``red shape''}) or color-shape combination (e.g., \textit{``red square''}). So the statement \textit{``A square is red''} and \textit{``A circle is blue''} are considered the same, while \textit{``A shape is red''} is different.

\begin{figure}[ht]
    \centering
    \includegraphics[width=0.45\textwidth]{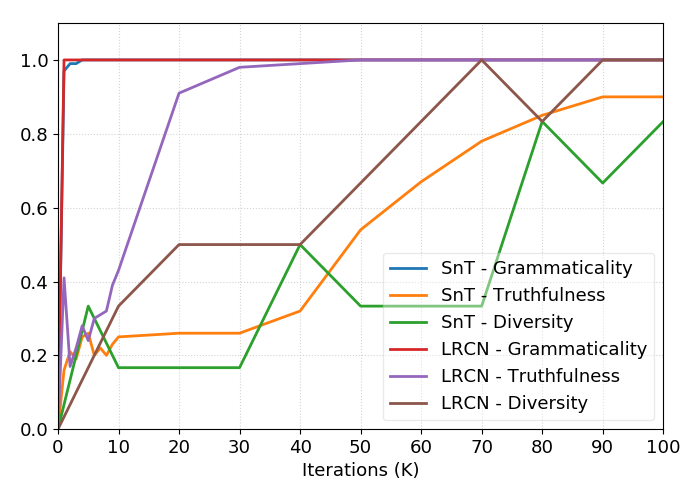}
    \centering
    \caption{Performance comparison of the Show\&Tell model and the LRCN\textsubscript{1u} model on \texttt{Existential-MultiShapes}. \textit{SnT} represents the Show\&Tell model while \textit{LRCN} represents the LRCN\textsubscript{1u} model. \textit{Grammaticality}, \textit{Truthfulness} and \textit{Diversity} refer to the grammaticality ratio, the truthfulness ratio and the diversity ratio of generated captions, respectively.}
    \label{fig:snt}
\end{figure}

\begin{figure*}[ht]
    \centering
    \hspace{-1cm}
    \begin{subfigure}[t]{0.30\textwidth}
        \centering
        \includegraphics[width=2.2in]{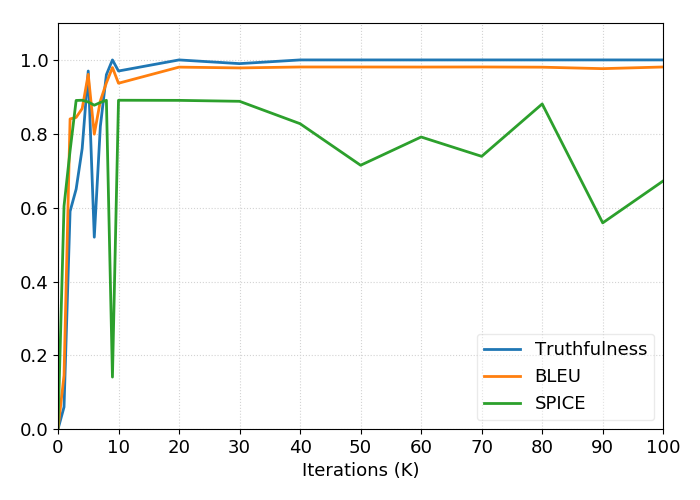}
        \caption{Existential-OneShape}
        \label{fig:existential-oneshape-metrics}
    \end{subfigure}
    \hspace{0.5cm}
    \begin{subfigure}[t]{0.30\textwidth}
        \centering
        \includegraphics[width=2.2in]{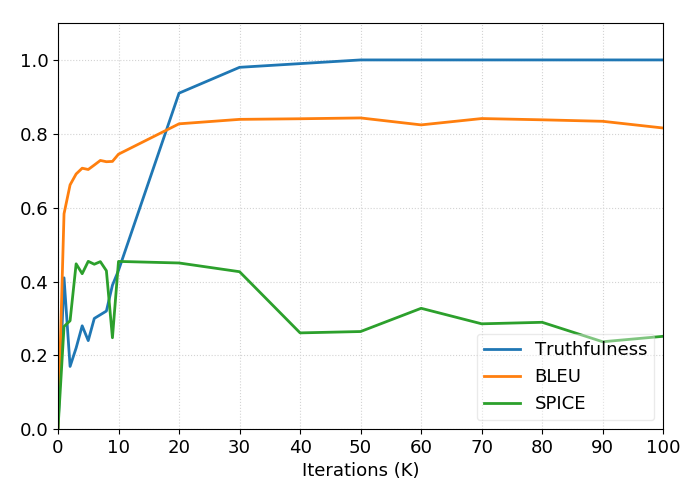}
        \caption{Existential-MultiShapes}
        \label{fig:existential-full_colfree-metrics}
    \end{subfigure}
    \hspace{0.5cm}
    \begin{subfigure}[t]{0.30\textwidth}
        \centering
        \includegraphics[width=2.2in]{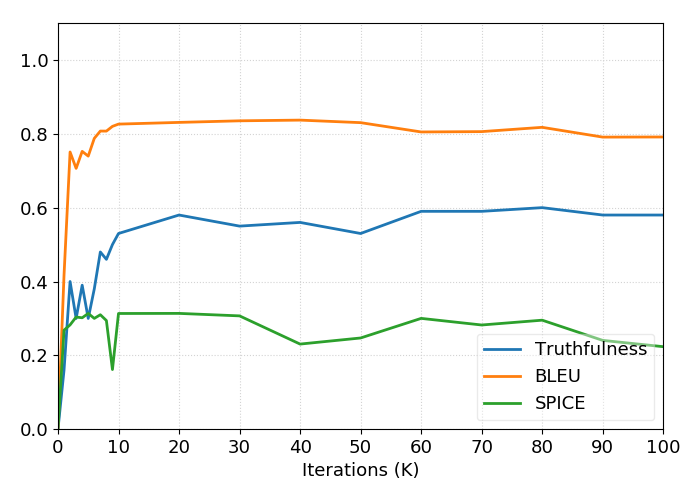}
        \caption{Spatial-MultiShapes}
        \label{fig:relational-spatial_reduced_colfree-metrics}
    \end{subfigure}
    \caption{Learning curves for LRCN\textsubscript{1u} on \texttt{Existential-OneShape}, \texttt{Existential-MultiShapes} and \texttt{Spatial-MultiShapes}. \textit{Truthfulness} refers to the ratio of generated captions that are grammatical and agree with ground-truth world models. \textit{BLEU} and \textit{SPICE} denote average BLEU-4 scores and average SPICE scores across the test split, respectively.}
    \label{fig:metrics}
\end{figure*}

\section{Experimental Setup}

\subsection{Datasets}

We develop a variety of ShapeWorldICE datasets, with a similar idea to the ``skill tasks'' in the bAbI framework \cite{weston2015towards}. Table \ref{tab:shapeworld} gives an overview for different ShapeWorldICE datasets we use in this paper. We consider three different types of captioning tasks, each of which focuses on a distinct aspect of reasoning abilities. Existential descriptions examine whether a certain object is present in an image. Spatial descriptions identify spatial relationships among visual objects. Quantification descriptions involve count-based and ratio-based statements, with an explicit focus on inspecting models for their counting ability. We develop two variants for each type of dataset to enable different levels of visual complexity or specific aspects of the same reasoning type. All the training and test captions sampled in this work are in English.

Each dataset variant consists of around 200k training instances and 4,096 validation instances, plus 4,096 test instances. Each training instance consists of an image and a reference caption. At test time, only the test images are available to the evaluated models. Underlying world models are kept from the models and are used for later GTD evaluation. For each test instance, we sample ten reference captions of the same distribution as the training captions to enable the comparison of our proposed metrics to BLEU and SPICE. We fine-tune our model hyperparameters based on the performance on the validation set. All reported results are measured on the test split with the parameters yielding the best validation performance.

\subsection{Models}

We experiment with two image captioning models: the Show\&Tell model \cite{vinyals2015show} and the LRCN\textsubscript{1u} model \cite{donahue2015long}. Both models follow the basic encoder-decoder architecture design that uses a CNN encoder to condense the visual information into an image embedding, which in turn conditions an LSTM decoder to generate a natural language caption. The main difference between the two models is the way they condition the decoder. The Show\&Tell model feeds the image embedding as the \textit{``predecessor word embedding''} to the first produced word, while the LRCN\textsubscript{1u} model concatenates the image features with the embedded previous word as the input to the sequence model at each time step.

We follow the common practice in image captioning to use a CNN component pretrained on object detection and fine-tune its parameters on the image captioning task. The encoder and decoder components are jointly optimized with respect to the standard cross-entropy sequence loss on the respective ShapeWorldICE dataset. For all our experiments, we train models end-to-end for a fixed number of 100k iterations with a batch size of 64. We use Adam optimization \cite{kingma2014adam} with a learning rate of 0.001. Word embeddings are randomly initialized and jointly trained during the training.

\section{Results}
\label{sec:results}

We train and evaluate the Show\&Tell and LRCN\textsubscript{1u} models on the ShapeWorldICE datasets. Here we discuss in detail the diagnostic results of these experiments. During training, we periodically record model output on the test images, to be able to analyze the development of our evaluation metrics throughout the process. We also compute BLEU-4 scores and SPICE scores of generated captions for comparison, using 10 reference captions per test image. 

\textbf{LRCN\textsubscript{1u} exhibits clearly superior performance in terms of truthfulness.}
We start off by comparing performance of the Show\&Tell model and the LRCN\textsubscript{1u} model, see Figure \ref{fig:snt}. While both models learn to produce grammatical sentences early on, it can be seen that LRCN\textsubscript{1u} is clearly superior in terms of truthfulness, achieving 100\% halfway through training, whereas Show\&Tell only slowly reaches around 90\% by the end of 100k iterations. This indicates that incorporating visual features at every generation step is beneficial for producing true captions. The diversity ratios of captions generated by two models both increase substantially as the training progresses, with LRCN\textsubscript{1u} exhibiting a slightly greater caption diversity at the end of training.

We observed similar results on other ShapeWorldICE datasets that we experimented with, validating the superiority of LRCN\textsubscript{1u} over Show\&Tell on ShapeWorldICE. Consequently, we decided to focus on the LRCN\textsubscript{1u} architecture in subsequent evaluations, where we report detailed results with respect to the GTD framework on a variety of datasets.

\textbf{Correlation between the BLEU/SPICE scores and the ground truth.} From the learning curves shown in Figure \ref{fig:metrics}, we find low or no correlation between the BLEU/SPICE scores and caption truthfulness. 

On \texttt{Existential-OneShape}, the BLEU curve follows the trend of the truthfulness curve in general, indicating that BLEU is able to capture caption truthfulness well in this simple scenario. However, while BLEU reports equivalent model performance on \texttt{Existential-MultiShapes} and \texttt{Spatial-MultiShapes}, the truthfulness metric demonstrates very different results. The BLEU score for generated \texttt{Existential-MultiShapes} captions increases rapidly at the beginning of training and then plateaus despite the continuous increase in truthfulness ratio. Captions generated on \texttt{Spatial-MultiShapes} attain a relatively high BLEU score from an early stage of training, but exhibit low agreement ($<$0.6 truthfulness ratio) with ground-truth visual scenes. In the case of \texttt{Spatial-MultiShapes}, spatial descriptors for two objects are chosen from a fixed set (\textit{``above''}, \textit{``below''}, \textit{``to the left of''} and \textit{``to the right of''}). It is very likely for a generated spatial descriptor to match one of the descriptors mentioned in reference captions. In this particular case, the model is apt to infer a caption which has high n-gram overlaps with reference captions, resulting in a relatively high BLEU score. Thus an increased BLEU score does not necessarily indicate improved performance.

While the truthfulness and BLEU scores in Figure \ref{fig:existential-oneshape-metrics} both increase rapidly early on and then stay stable at a high rate after training for 20k iterations, the SPICE curve instead shows a downward trend in the later stage of training. We examined the output SPICE score for each test instance. SPICE reports a precision score of 1.0 for most test instances after 20k iterations, which is consistent with the truthfulness and BLEU scores. However, SPICE forms the reference scene graph as the union of the scene graphs extracted from individual reference captions, thus introducing redundancies. SPICE uses the F1 score of scene graph matching between the candidate and reference and hence is lowered by imperfect recall.

Comparing SPICE curves for three datasets shown in Figure \ref{fig:existential-oneshape-metrics}-\ref{fig:relational-spatial_reduced_colfree-metrics}, they suggest an increase in task complexity, but they do not reflect the successively closing gap of caption truthfulness scores between two \texttt{Existential} datasets, or the substantial difference in caption truthfulness between captions on \texttt{Existential-MultiShapes} and \texttt{Spatial-MultiShapes}.


In the remainder of the paper we discuss in detail the diagnostic results of the LRCN\textsubscript{1u} model demonstrated by the GTD evaluation framework.

\begin{figure}[ht]
    \centering
    \includegraphics[width=0.45\textwidth]{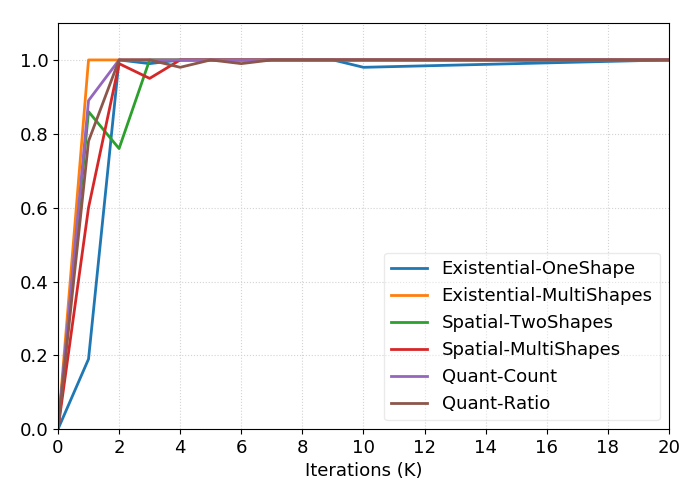}
    \centering
    \caption{Ratio of grammatical sentences produced by LRCN\textsubscript{1u} for different ShapeWorldICE datasets in the first 20k training iterations (stays at 100\% afterwards).}
    \label{fig:grammaticality}
\end{figure}

\textbf{Perfect grammaticality for all caption types.}
As shown in Figure \ref{fig:grammaticality}, generated captions for all types of ShapeWorldICE datasets attain quasi-perfect grammaticality scores in fewer than 5,000 iterations, suggesting that the model quickly learns to generate grammatically well-formed sentences.

\begin{figure}[ht]
    \centering
    \includegraphics[width=0.45\textwidth]{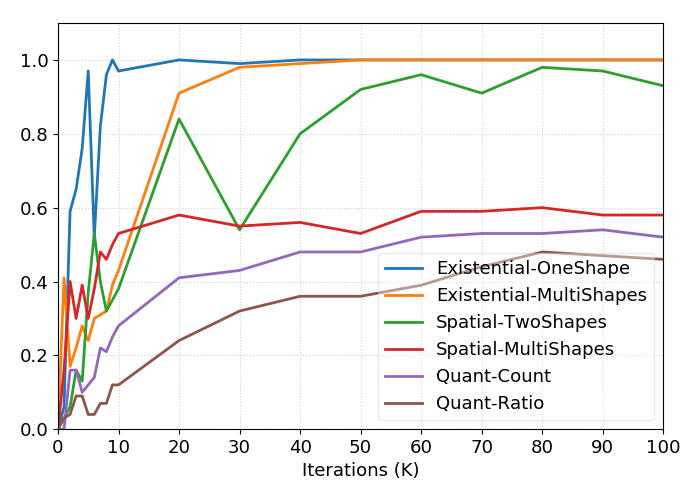}
    \centering
    \caption{Truthfulness ratio of sentences produced by LRCN\textsubscript{1u} for different ShapeWorldICE datasets.}
    \label{fig:truthfulness}
\end{figure}

\textbf{Failure to learn complex spatial relationships.} While CNNs can produce rich visual representations that can be used for a variety of vision tasks \cite{sermanet2013overfeat}, it remains an open question whether these condensed visual representations are rich enough for multimodal tasks that require higher-level abilities of scene understanding and visual reasoning. From Figure \ref{fig:truthfulness}, we can see that while the model performs rather well on \texttt{Existential} datasets, it exhibits a worse performance on \texttt{Spatial} data. The caption agreement ratio in the simple \texttt{Spatial-TwoShapes} scenario is relatively high, but drops significantly on \texttt{Spatial-MultiShapes}, demonstrating the deficiencies of the model in learning spatial relationships from complex visual scenes.

\textbf{The counting task is non-trivial.}
Counting has long been considered to be a challenging task in multimodal reasoning \cite{antol2015vqa,jabri2016revisiting}. To explore how well the LRCN\textsubscript{1u} model copes with counting tasks, we generated two \texttt{Quantification} datasets. The \texttt{Quant-Count} captions describe the number of objects with certain attributes that appear in an image (\textit{e.g. ``Exactly four shapes are crosses''}), while the \texttt{Quant-Ratio} captions describe the ratio of certain objects (\textit{e.g. ``A third of the shapes are blue squares''}).

From Figure \ref{fig:truthfulness}, we notice that the LRCN\textsubscript{1u} model performs poorly on these datasets in terms of truthfulness, reflected in the 0.50 and 0.46 scores achieved by the model on the \texttt{Quant-Count} and \texttt{Quant-Ratio} tasks respectively. The learning curve for \texttt{Quant-Ratio} exhibits a more gradual rise as the training progresses, suggesting a greater complexity for the ratio-based task.

\begin{figure}[ht]
    \centering
    \includegraphics[width=0.45\textwidth]{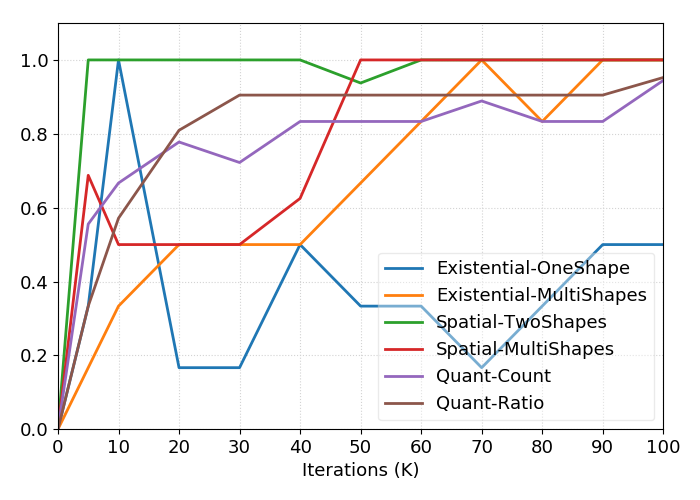}
    \caption{Diversity ratio of sentences produced by LRCN\textsubscript{1u} on different ShapeWorldICE datasets.}
    \label{fig:diversity}
\end{figure}

\textbf{Caption diversity benefits from varied language constructions in the training data.} The diversity ratios of generated captions for different ShapeWorldICE datasets are illustrated in Figure \ref{fig:diversity}. We can see that the diversity of inferred captions is largely sensitive to the caption variability in the dataset itself. For simple datasets (such as \texttt{Existential-OneShape}) where language constructions in the training set are less diverse, the output captions tend to have uniform sentence structures. The high diversity ratios of generated \texttt{Spatial} and \texttt{Quantification} captions suggest that caption diversity benefits from heterogeneous language constructions in complex datasets.

\section{Discussions and Conclusions}
\label{sec:conclusion}
Evaluation metrics are required as a proxy for performance in real applications. As such, they should, as far as possible, allow measurement of fundamental aspects of the performance of models on tasks. In this work, we propose the GTD evaluation framework as a supplement to standard image captioning evaluation which explicitly focuses on grammaticality, truthfulness and diversity. We developed the ShapeWorldICE evaluation suite to allow in-depth and fine-grained inspection of model behaviors. We have empirically verified that GTD captures different aspects of performance to existing metrics by evaluating image captioning models on the ShapeWorldICE suite. We hope that this framework will shed light on important aspects of model behaviour and that this will help guide future research efforts.

While performing the evaluation experiments on the LRCN\textsubscript{1u} model, we noticed that caption agreement does not always improve as the training loss decreases. Ideally, the training objective should be in accordance with how a model is eventually evaluated. In future work, we plan to investigate the feasibility of deliberately encoding the GTD signal in the training process, for instance, by implementing a GTD-aware loss. We also plan to extend the existing ShapeWorldICE benchmark to include more linguistic constructions (such as relative clauses, compound sentences and coreference). By doing so, we hope to reveal how well existing image captioning models cope with complex generation tasks.

\section{Acknowledgments}
We thank the anonymous reviewers for their constructive feedback. HX is grateful for being supported by the CSC Cambridge Scholarship. TS is supported in part by the EPSRC Centre for Doctoral Training in Data Science, funded by the EPSRC (grant EP/L016427/1) and the University of Edinburgh. AK is grateful for being supported by a Qualcomm Research Studentship and an EPSRC Doctoral Training Studentship.

\bibliographystyle{aaai}
\bibliography{aaai20}

\end{document}